\documentclass[letterpaper]{article} % DO NOT CHANGE THIS
\usepackage[]{aaai24}  % DO NOT CHANGE THIS
\usepackage{times}  % DO NOT CHANGE THIS
\usepackage{helvet}  % DO NOT CHANGE THIS
\usepackage{courier}  % DO NOT CHANGE THIS
\usepackage[hyphens]{url}  % DO NOT CHANGE THIS
\usepackage{graphicx} % DO NOT CHANGE THIS
\usepackage{array}
\usepackage{natbib}  % DO NOT CHANGE THIS AND DO NOT ADD ANY OPTIONS TO IT
\usepackage{caption} % DO NOT CHANGE THIS AND DO NOT ADD ANY OPTIONS TO IT
\usepackage{algorithm}
\usepackage{algorithmic}
\usepackage{multirow}

\usepackage{amsfonts,amsmath}

\usepackage{newfloat}
\usepackage{listings}
\DeclareCaptionStyle{ruled}{labelfont=normalfont,labelsep=colon,strut=off} % DO NOT CHANGE THIS
\floatstyle{ruled}
\newfloat{listing}{tb}{lst}{}
\floatname{listing}{Listing}
\title{Interpreting Time Series Transformer Models and Sensitivity Analysis of Population Age Groups to COVID-19 Infections}
\author{Md Khairul Islam \textsuperscript{\rm 1}, Tyler Valentine \textsuperscript{\rm 2}, Timothy Joowon Sue \textsuperscript{\rm 1}, Ayush Karmacharya \textsuperscript{\rm 1},   Luke Neil Benham 
\textsuperscript{\rm 1}, Zhengguang Wang \textsuperscript{\rm 1}, Kingsley Kim \textsuperscript{\rm 1}, Judy Fox \textsuperscript{\rm 1, 2}}
\affiliations{
\textsuperscript{\rm 1} Computer Science, University of Virginia, USA. \\
\textsuperscript{\rm 2} School of Data Science, University of Virginia, USA. \\
Email: \{mi3se, xje4cy, und2yw, psb7wm, lnb6grp, zw4re, bjb3az, cwk9mp\}@virginia.edu
}

\begin{document}
\maketitle
\begin{abstract}
Interpreting deep learning time series models is crucial in understanding the model's behavior and learning patterns from raw data for real-time decision-making. However, the complexity inherent in transformer-based time series models poses challenges in explaining the impact of individual features on predictions. In this study, we leverage recent local interpretation methods to interpret state-of-the-art time series models. To use real-world datasets, we collected three years of daily case data for 3,142 US counties. Firstly, we compare six transformer-based models and choose the best prediction model for COVID-19 infection. Using 13 input features from the last two weeks, we can predict the cases for the next two weeks. Secondly, we present an innovative way to evaluate the prediction sensitivity to 8 population age groups over highly dynamic multivariate infection data. Thirdly, we compare our proposed perturbation-based interpretation method with related work, including a total of eight local interpretation methods. Finally, we apply our framework to traffic and electricity datasets, demonstrating that our approach is generic and can be applied to other time-series domains. 

\end{abstract}

\section{Introduction}
As the promise of deep learning models in critical domains \cite{zhao2023interpretation} such as health, finance, and social science, ensuring the interpretability of these methods becomes essential for maintaining AI transparency and the reliability of model decisions \cite{Amann2020ExplainabilityFA}.

The interpretability methods evaluate different factors contributing to the model decisions \cite{rojat2021explainable}, reveal incompleteness in the problem formalization, and improve our scientific understanding \cite{doshi2017towards}. Explaining time series models in a meaningful way is challenging due to their dynamic nature. Much research on time series models has focused on interpreting classification tasks or using simple models. It is desirable to understand how to use these interpretation methods in state-of-the-art time series models while still achieving the best performance. 

In summary, the main contributions of our research are:
\begin{itemize}
\item Use the highly dynamic data in a multivariate, multi-horizon, and multi-modal setting with state-of-the-art time series transformer models. COVID-19 is a recent pandemic taking millions of lives and causing many research efforts to forecast the infection spread using statistical learning, epidemiological, and machine learning models \cite{clement2021survey}. 
\item Collect around three years of COVID-19 data daily for 3,142 US counties. Each county contributes to one time series in the dataset (hence multi-time series). We use the last 14 days of data to predict the cases for the next 14 days. The best-performing model on the test set is later used for interpretation. This allows us to give a more granular analysis. 
\item Focus on local interpretation methods to show the contribution of each input feature to the prediction given an input sample, and benchmark our interpretation using eight recent methods. Then evaluate the interpreted attribution scores following the latest practices \cite{ozyegen2022evaluation}. 
\item Propose an innovative way to evaluate sensitivities of age group features using infection by age group. We employ a black-box interpretation method in our approach, which applies to various models and time series datasets.
\end{itemize}
The rest of the sections are organized as follows: Section \ref{sec:background_related_works} describes the background and related work. Section \ref{sec:problem_statement} defines the problem statement of both forecasting and interpretation tasks. Section \ref{sec:dataset} describes the data collection and pre-processing steps. Section \ref{sec:exp_setup} summarizes the experimental setup, training steps, and interpretation without ground truth for all of our three datasets.  Section \ref{sec:ground_truth} discusses evaluating age group sensitivity with ground truth for our COVID-19 dataset. Sections \ref{sec:discussion} and \ref{sec:conclusion} examine additional aspects of our approach and conclude our work. 

\section{Background and Related Work \label{sec:background_related_works}}
In this study, we focus on local interpretation methods consisting of both time series interpretation and the prediction of COVID-19 infection. The following sections contain the related terminologies and recent works on this topic.

\subsection{Background Terminologies}
Interpretation methods are either: 1)\textbf{White box}: Using the model's inherent architecture to interpret model behavior (e.g. using attention weights). 2) \textbf{Black box}: Using only the input and output to determine the model's behavior. We use black box methods in this work since they are model-agnostic and more applicable. 

Based on application scope, interpretability methods are either: 1) \textbf{Global}: Explains the entire behavior of the model 2) \textbf{Local}: Explains the reasons behind a specific model decision on an input instance \cite{rojat2021explainable}. We focus on local interpretation methods in this work, to provide a granular analysis of the model's behavior.

\textbf{Interpretation methods} aim to quantify the relevance of input features to model output. This helps identify key features influencing the model's decision. Evaluating these importance scores in practice is difficult due to the lack of ground truth for interpretation. However, existing studies \cite{rojat2021explainable, ismail2020benchmarking} perturb the top features based on these interpreted importance scores and recalculate how much that impacts the model's output to evaluate interpretation quantitatively (Section \ref{sec:eval_without_truth}). 

\textbf{Sensitivity analysis} is one type of perturbation-based technique to interpret a model's behavior. Morris method \cite{morris1991factorial} is a sensitivity analysis method that defines the sensitivity of a model input as the ratio of the change in an output variable to the change in an input feature. We expanded the Morris method to temporal data using the SALib library \cite{Iwanaga2022}. We use the `mu\_star`  as the feature importance score, as it is more reliable. A higher `mu\_star` indicates higher sensitivity.

\subsection{Related Work}

In this study, we focus on local interpretation methods consisting of both time series interpretation and the prediction of COVID-19 infection. 

\subsubsection{Time Series Interpretation}
A wide range of interpretation methods has been proposed in the literature \citep{rojat2021explainable, turbe2023evaluation}. Including \textit{gradient based methods} such as Integrated Gradients \citep{Sundararajan2017AxiomaticAF}, GradientSHAP \citep{Erion2019LearningEM} which uses the gradient of the model predictions to input features to generate importance scores. \textit{Feature removal based} methods such as Feature Occlusion \citep{Zeiler2013VisualizingAU}, Feature Ablation \citep{Suresh2017ClinicalIP}, and Sensitivity Analysis \citep{morris1991factorial} replace a feature or a set of features from the input using some fixed baselines or generated samples and measure the importance based on the model output change. 

These methods have been popular and used in time series datasets \cite{ozyegen2022evaluation, zhao2023interpretation, turbe2023evaluation}. \citet{ismail2020benchmarking} standardized the evaluation of local interpretations when no interpretation ground truth is present. Most prior works on time series interpretation focus on classification \cite{turbe2023evaluation, ismail2020benchmarking, rojat2021explainable}. \citet{ozyegen2022evaluation} proposed a novel evaluation metric for local interpretation in time series regression tasks using real-world datasets. 

One key limitation of the prior works is using baseline models or synthetic datasets, which doesn't reflect the practical case where we want to use state-of-the-art models with complex real-world datasets. We address this limitation by incorporating recent time series models in our work.
% \textit{Model based saliency} methods such as \citep{Choi2016RETAINAI, Song2017AttendAD, Xu2018RAIMRA, Kaji2019AnAB, lim2021temporal} use the model architecture e.g. attention layers, to generate importance scores.

\subsubsection{Interpreting COVID-19 Infection}
DeepCOVID \cite{rodriguez2021deepcovid} utilized RNN with auto-regressive inputs to predict COVID-19 cases. Then recursively eliminating input signals to quantify the model output deviation without those signals and use that to rank the signal importance. DeepCOVIDNet \cite{ramchandani2020deepcovidnet} classified infected regions with high, medium, and low case growth, then interpreted using Feature Occlusion on part of the training data. 

COVID-EENet \cite{kim2022covid} interpreted the economic impact of COVID-19 on local businesses. Self-Adaptive Forecasting \cite{arik2022self} used the model's attention weights to interpret state-level death forecasts. However, this is model-dependent and can't be applied to models without the attention mechanism. 

One key limitation of these works is not comparing their interpretation performance with other interpretation methods. In this work, we address this challenge and bridge the gap by comparing eight recent local interpretation methods.

\section{Problem Statement \label{sec:problem_statement}}
We consider a multivariate multi-horizon time series setting with length $T$, the number of input features $J$, and total $N$ instances. $X_{j,t} \in \mathbb{R}^{J \times T}$ is the input feature $j$ at time $t \in \{0, \cdots, T-1\}$. We use past information within a fixed look-back window $L$, to forecast for the next $\tau_{max}$ time steps. The target output at time $t$ is $y_t$. Hence our black-box model $f$ can be defined as $\hat{y}_{t} = f(X_t)$ where,

\begin{equation}
\begin{aligned}
    % \hat{y}_{t} & = f(X_t)  \\ \text{where, } 
 X_t &= x_{t-(L-1):t} \\
 &= [x_{t-(L-1)}, x_{t-(L-2)}, \cdots, x_t] \\
&= \{ x_{j, l, t}\}, ~ j \in \{1, \cdots, J\}, ~ l \in \{1, \cdots, L\}
\end{aligned}
\end{equation}

$\hat{y}_{t}$ is the forecast at $\tau \in \{1, \cdots, \tau_{max}\}$ time steps in the future.  $ X_t$ is the input slice at time $t$ of length $L$. 
An individual covariate at position $(n, l)$ in the full covariate matrix at time step $t$ is denoted as $x_{j, l, t}$. 

For interpretation, our target is to construct the importance matrix $\phi_t = \{ \phi_{j, l, t} \}$ for each output $o \in O$ and prediction horizon $\tau \in \{1, \cdots, \tau_{max}\}$. So this is a matrix of size $O \times \tau_{max} \times J \times L$. We find the relevance of the feature $x_{j, l, t}$ by masking it in the input matrix $X_t$ and output change from the model,
\begin{equation}
    \phi_{j, l, t} = | (f(X_t) - f(X_t ~\text{\textbackslash}~ x_{j, l, t})|
\end{equation}
where $X_t ~\text{\textbackslash}~ x_{j, l, t}$ is the feature matrix achieved after masking entry $x_{j, l, t}$.

\subsection{Methodology for Local Interpretation of Time-Series\label{sec:eval_without_truth}}

A major challenge in evaluating interpretation is the lack of interpretation of ground truth. 
We use the following quantitative analysis steps \cite{ozyegen2022evaluation} to \textbf{perform \textit{local interpretation evaluation} in the absence of ground truth}:

\begin{enumerate}
    \item Sort relevance scores $R(X)$ returned by the interpretation method so that $R_e(X_{i, t})$ is the $e^{th}$ element in the ordered set $\{R_e(x_{i, t})_{e=1}^{L \times N}\}$. Here $L$ is the look-back window and $N$ is the number of features. 
    \item Find top $k\% ~ (\text{we used, } k \in \{5, 7.5, 10, 15\})$ entries in this set, where $\mathbf{R}(x_{i, t}) \in \{\mathbf{R}_e(x_{i, t})\}^k_{e=1}$. 
    \item Mask these top features or every other feature. 
    \item Calculate the change in the model's output to the original output using the mean absolute error (MAE) metric following \cite{ozyegen2022evaluation}. 
\end{enumerate}

\citet{deyoung2019eraser} propose the \textit{comprehensiveness} and \textit{sufficiency} metrics to measure the faithfulness of the interpretations. \textit{Comprehensiveness} defines whether all features needed to make a prediction were selected, measured by calculating the output change after masking the top important features. Intuitively, the model should be less confident in its prediction afterward. \textit{Sufficiency} defines whether the features selected as important contain enough information to make the prediction. This is measured by masking any other feature except the top important features. The smaller the output change, the more sufficient the selected features are.

In summary, \textbf{the higher the comprehensiveness loss and the lower the sufficiency loss the better}. We define the set of top $k\%$ relevant features selected by the interpretation method for the $i$-th input $X_i$ as $X_{i, 1:k}$, input after removing those features $X_{i, ~\text{\textbackslash} 1:k}$. Then for our model $f()$ we can describe comprehensiveness and sufficiency as:
\begin{equation}
\begin{aligned}
    & \text{Comprehensiveness} &= |f(X_i) - f(X_{i, ~\text{\textbackslash} 1:k} ) | \\
    & \text{Sufficiency} &= |f(X_i) - f(X_{i, 1:k}) |
\end{aligned} 
\end{equation}
For $K$ bins of top $k\%$ features (we use top 5\%, and 10\% features, hence $K=2$.), the aggregated comprehensiveness score is referred to as the "Area Over the Perturbation Curve for Regression" or AOPCR \cite{ozyegen2022evaluation}. 
\begin{equation}
    AOPCR = \frac{1}{K \times \tau_{max}} \sum_\tau^{\tau_{max}} \sum_k^K  |f(X_i)_\tau - f(X_{i,  ~\text{\textbackslash} 1:k})_\tau) |
\end{equation}
We calculate the AOPCR for sufficiency similarly after replacing $X_{i,  ~\text{\textbackslash} 1:k}$ with $X_{i, 1:k}$.  Table \ref{tab:int_results_no_ground_truth} presents the AOPCR results from the interpretation methods. It shows our implementation of the Morris Sensitivity method performing the best in most cases (3 out of 4). 

\subsection{Interpretation Methods\label{sec:interpretation}}
In this section, we describe our interpretation methods and how to evaluate the interpretation performance across different methods. The interpretation is done using the FEDformer model on the test set. However, the approach is generic and model-agnostic. Therefore it can be used for other time series models.

We use the following recent methods to perform black-box local interpretation analysis on the target mode: 
\begin{enumerate}
    \item \textit{\textbf{Feature Ablation (FA):}} Computes \cite{Suresh2017ClinicalIP}  attribution as the difference in output after replacing each feature with a baseline. 
    \item \textit{\textbf{Feature Permutation (FP):}} Permutes the  \citep{molnar2020interpretable}  the input feature values within a batch, and computes the difference between original and shuffled outputs.
    \item \textit{\textbf{Morris Sensitivity (MS):}} Morris method \cite{morris1991factorial} calculates the model output change with respect to a $\delta$ change to the input value. We designed a temporal adaptation of this Morris method using the Sensitivity Analysis Library \cite{Iwanaga2022}. 
    \item \textit{\textbf{Feature Occlusion (FO):}} Replaces the input features with a counterfactual generated from a normal distribution \cite{Suresh2017ClinicalIP}.
    \item \textit{\textbf{Augmented Feature Occlusion (AFO):}} Augments the Feature Occlusion method by sampling counterfactuals from the bootstrapped distribution over each feature, avoiding generating out-of-distribution samples \cite{Suresh2017ClinicalIP}. 
    
    \item \textit{\textbf{Deep Lift (DL):}}  Deep Learning Important FeaTures \cite{shrikumar2017learning} method decomposes the output prediction of a neural network on a specific input by backpropagating the contributions of all neurons in the network to every feature of the input. 
    \item \textit{\textbf{Integrated Gradients (IG):}} Assigns This method \cite{Sundararajan2017AxiomaticAF} assigns an importance score to each input feature by approximating the integral of gradients of the model’s output to the inputs along the path (straight line) from given baselines/references to inputs.
    \item \textit{\textbf{Gradient Shap (GS):}} Uses the gradient of the model predictions to input features to generate importance scores \cite{Erion2019LearningEM}.
    % \item \textit{Lime:} Trains an interpretable surrogate model \cite{ribeiro2016should} by sampling points around a specified input example and using model evaluations at these points to train a simpler interpretable ‘surrogate’ moCOIVD-19such as a linear model.
\end{enumerate}

\section{Datasets \label{sec:dataset}}
In this section, we describe three datasets used in our experiments and their respective initial data processing steps. We compile a new \textbf{COVID-19 dataset} dataset, on which we perform both our proposed window-based time series interpretation and sensitivity analysis. Furthermore, to answer our research question: \textbf{Is our proposed window-based time series interpretation framework applicable to other models and datasets?}
Our proposed workflow is model-agnostic and generic, hence can be applied to any other time series models and datasets to interpret input-output relevance. We answer this by choosing two well known time series datasets: \textbf{Electricity} and \textbf{Traffic}. 

\begin{itemize}
\item \textit{\textbf{COVID-19 dataset}}: Our data is collected from multiple public sources from March 1, 2020, to Dec 29, 2022 (around 3 years) for each of 3,142 US counties. Additionally, we collected weekly COVID-19 cases by age groups from \cite{weekly_cases} to evaluate the age group sensitivity interpretation. These age groups are categorized by the US county population statistics \cite{pop2020} and cases by age groups from CDC \cite{weekly_cases}. 

We removed outliers from the data using the following thresholds:
\begin{equation}
\begin{aligned}
\text{lower} = Q1-(7.5*IQR) \\
\text{upper} = Q3+(7.5*IQR)
\end{aligned}
\end{equation}

where $Q1$ and $Q3$ represent the first and third percentiles on a weekly moving average basis, and $IQR$ is the interquartile range. We linearly interpolated the missing values and standard normalized the features before training the model. The model uses the previous 2 weeks of data to predict cases for the next 2 weeks.

\newcolumntype{P}[1]{>{\centering\arraybackslash}p{#1}}
\newcolumntype{M}[1]{>{\centering\arraybackslash}m{#1}}
\begin{table}[htbp]
    \centering
    \begin{tabular}{|M{2cm}|M{3.1cm}|M{2.2cm}|} \hline
        \textbf{Feature}  & \textbf{Description} & \textbf{Type}  \\ \hline
         Age groups & \% of people in each of the 8 age groups & Static  \\ \hline
         Vaccination & \% of fully vaccinated population & Dynamic \\ \hline
         Cases & Past infection cases & Dynamic \\ \hline
         Month & Timestamps & Known Future  \\\hline
         Day in month & Timestamps & Known Future  \\\hline
         Day in week & Timestamps & Known Future  \\\hline
         Cases & Future infection cases & Target \\ \hline
    \end{tabular}
    \caption{Description of the dataset. Data is collected for each of the 3,142 US counties.}
\end{table}

\item \textit{\textbf{Electricity dataset}}: The UCI Electricity Load Diagrams dataset contains the electricity consumption of 321 customers from 2012 to 2014. They are aggregated to an hourly level and normalized. We use the past 96 hours of inputs to forecast for the next 24 hours. We also added four time-encoded features: month, day of the month, day of the week, and hour. 
Following \cite{zhang2022crossformer} we use the record of customer `MT\_321` as the time series of interest. 
\begin{table}[htbp]
    \centering
    \begin{tabular}{|M{2cm}|M{3.1cm}|M{2.2cm}|} \hline
        \textbf{Feature}  & \textbf{Description} & \textbf{Type}  \\ \hline
         Consumption  & Past electricity consumption  & Dynamic \\ \hline
         Month & Timestamps & Known Future  \\\hline
         Day in month & Timestamps & Known Future  \\\hline
         Day in week & Timestamps & Known Future  \\\hline
         Hour & Timestamps & Known Future \\\hline
         Consumption & Future electricity consumption & Target \\ \hline
    \end{tabular}
    \caption{Description of Electricity dataset.}
\end{table}

\item \textit{\textbf{Traffic dataset}}: The UCI PEM-SF Traffic dataset describes the occupancy rate (with $y_t \in [0, 1]$) of 440 SF Bay Area freeways from 2015 to 2016. We perform the same data processing steps with the Electricity dataset. Following \cite{zhang2022crossformer} we use the record of 821st station user as the time series of interest. 
\begin{table}[htbp]
    \centering
    \begin{tabular}{|M{2cm}|M{3.1cm}|M{2.2cm}|} \hline
        \textbf{Feature}  & \textbf{Description} & \textbf{Type}  \\ \hline
         Occupancy rate  & Past road occupancy rate  & Dynamic \\ \hline
         Month & Timestamps & Known Future  \\\hline
         Day in month & Timestamps & Known Future  \\\hline
         Day in week & Timestamps & Known Future  \\\hline
         Hour & Timestamps & Known Future \\\hline
         Occupancy rate & Future road occupancy rate & Target \\ \hline
    \end{tabular}
    \caption{Description of Traffic dataset.}
\end{table}
\end{itemize}

%Both have been used as a benchmark in time series forecasting works %\citep{wu2023timesnet, zhang2022crossformer, lim2021temporal, %ozyegen2022evaluation}. They are aggregated to an hourly level and normalized. We %use the past 96 hours of inputs to forecast for the next 24 hours. We also added %four time-encoded features: month, day of the month, hour, and day of the week. %The model uses the past target values and these time-encoded features as inputs.
%\begin{itemize}
%\item\textbf{Electricity :} The UCI Electricity Load Diagrams dataset contains the %electricity consumption of 321 customers from 2012 to 2014. Following %\cite{zhang2022crossformer} we use the `MT\_321` as the target variable. 
%\item\textbf{Traffic :} The UCI PEM-SF Traffic dataset describes the occupancy %rate (with $y_t \in [0, 1]$) of 440 SF Bay Area freeways from 2015 to 2016. %Following \cite{zhang2022crossformer} we use the 821st user as the target variable.
%\end{itemize}

\section{Experiment Setup}
\label{sec:exp_setup}

This section describes our experimental setup for preparing the best time series model for the infection forecasting task at the daily US county level. In the next section, we describe how we interpret this model using black-box interpretation methods. All of our experiments were done in a remote HPC server with NVIDIA V100 GPU and 32 GB memory. 

% \begin{table*}[ht]
% \centering

% %\small
% \begin{tabular}{lllm{16.5em}}

% \hline
% Feature   &Variable  &Type    &Description  \\
%  \hline
% \multirow{8}{*}{Age groups} & Under5  & Static & \% of people under 5 years old  \\
%  &Age517 &  Static  & \% of people between 5 and 17 years old   \\
% & Age1829 & Static & \% of people between 18 and 29 years old \\
% & Age3039 & Static & \% of people between 30 and 39 years old  \\
% & Age4049 & Static & \% of people between 40 and 49 years old  \\
% & Age5064 & Static & \% of people between 50 and 64 years old  \\
% & Age6574 & Static & \% of people between 65 and 74 years old  \\
% & Age75Plus & Static & \% of people above 75 years old  \\
% \hline
% \multirow{1}{2em}[-0.1ex]{VaccinatiCOVID-19 VaccinationFull  & Dynamic & \% of fully vaccinated population \\
% \hline
% \multirow{1}{*}[-0.1ex]{Past Cases} & Cases  & Dynamic & Past Infection Cases\\
% \hline
% \multirow{1}{*}[-0.1ex]{Month, day of the month, day of the week} & Date  & Known Future & Timestamps \\
% \hline
% \multirow{1}{2em}[-0.1ex]{Cases} & Cases  & Target & Current Infection Cases \\
% \hline

% \end{tabular}
% \caption{Description of the dataset. Data is collected for each of the 3,142 US counties.}
% \label{tab:dataset}
% \end{table*}

\subsection{COVID-19 Data Preprocessing}

The data set was split into training, validation, and testing sets. The training set includes March 1, 2020, through November 27, 2021. The immediate next 2 weeks are used as the validation set and the next 2 weeks after that are used as the test set. The best model checkpointed by the validation set is loaded later for testing. We use additional data after this period for deployment benchmark in Section \ref{sec:interpretation}.

%\section{Age Group Cases Statistics}

% \begin{table}[htbp]
%     \centering
%     \begin{tabular}{|c|c|c|c|c|} \hline
%         \multirow{2}{*}{\textbf{Age Feature}} & \multicolumn{2}{|c|}{\textbf{Normalized}} & \multicolumn{2}{|c|}{\textbf{Rank}} \\ \cline{2-5}
%         & \textbf{Mean} & \textbf{Std.}  & \textbf{Mean} & \textbf{Std.} \\ \hline
%         UNDER5 & 0.033 & 0.014 & 7.858 & 0.350 \\
%         AGE517 & 0.118 & 0.050 & 4.757 & 1.828 \\
%         AGE1829 & 0.202 & 0.041 & 1.662 & 0.952 \\
%         AGE3039 & 0.164 & 0.012 & 2.980 & 0.540 \\
%         AGE4049 & 0.142 & 0.011 & 4.297 & 0.769 \\
%         AGE5064 & 0.194 & 0.028 & 1.986 & 0.911 \\
%         AGE6574 & 0.079 & 0.022 & 5.872 & 0.409 \\
%         AGE75PLUS & 0.067 & 0.031 & 6.588 & 1.223 \\
%         \hline
%     \end{tabular}
%     \caption{Statistics of normalized weekly COVID-19 cases by population age groups and ranks in the USA during our experiment period: 1 March 2020 to 31 Dec 2022.}
%     \label{tab:age_ground_truth_stats}
% \end{table}

% Table \ref{tab:age_ground_truth_stats} shows the statistics of the ground truth COVID-19 cases by age group. We present a summary of both normalized values (l1-norm of each week's cases from different age groups) and the rank of those normalized cases. AGE1829 generally stays at rank 1 in the weekly COVID-19 case count, where UNDER5 populations generally contribute the lowest numbers.

\subsection{Models and Parameters}

We use the following models for benchmarking: TimesNet \cite{wu2023timesnet}, PatchTST \cite{nie2023patchtst}, FEDformer \cite{zhou2022fedformer}, Autoformer \cite{wu2021autoformer}, Crossformer \cite{zhang2022crossformer}. We implement these models \footnote{https://github.com/UVA-MLSys/COVID-19-age-groups} using The Time-Series-Library \cite{TSlib}. 

The model hyper-parameters were chosen based on \cite{wu2023timesnet} and tuning. Table \ref{tab:parameters} reports the common parameters used by the selected models during the experiment. Full documentation can be found in our project repository.

  \begin{table}[htbp]
      \centering
      \begin{tabular}{|p{2.2cm}|c|p{1.8cm}|c|} \hline
            \textbf{Parameter} & \textbf{Value} & \textbf{Parameter} & \textbf{Value} \\ \hline
           learning rate & 1e-3  & loss & MSE \\
           batch size & 32 & dropout & 0.1 \\
           encoder layers & 2 & random seed & 7 \\
           decoder layers & 1 & hidden size & 64\\
           attention heads & 4 & label length & 7\\ \hline
      \end{tabular}
      \caption{Common model hyperparameters.}
      \label{tab:parameters}
  \end{table}

\subsection{Implementation Library}
We used the Captum \cite{kokhlikyan2020captum} and Time Interpret \cite{enguehard2023time} libraries to implement these interpretation methods. Except for the Morris Sensitivity \cite{morris1991factorial}, which was implemented using the Sensitivity Analysis Library \cite{Iwanaga2022}. The baselines to mask the input features were randomly generated from a normal distribution. Unlike \cite{enguehard2023time}, which runs the interpretation on CPU, our implementation modifies the prior libraries to work on GPU. For interpretation methods requiring a bootstrapped distribution for baseline generation, we used the training data as the distribution.  

\subsection{Prediction Results}

We use the following popular evaluation metrics for our regression task: Mean Absolute Error (MAE), Root Mean Squared Error (RMSE), Root Mean Squared Logarithmic Error (RMSLE), and Coefficient of determination ($R^2$-score). For all metrics, except $R^2$-score, the lower the error loss the better. For R$^2$-score, 1.0 is the best possible score and it can be negative if the model is arbitrarily worse. 

\begin{table}[htbp]
    \centering
    \begin{tabular}{|p{1.7cm}|cccc|} \hline
        \textbf{Model} & \textbf{MAE} & \textbf{RMSE}  & \textbf{RMSLE} & \textbf{R$^2$-score}\\ \hline
         Autoformer & 35.69 & 189.4 & 1.918 & 0.451\\
         FEDformer & \textbf{30.19} & \textbf{182.2} & \textbf{1.467} & \textbf{0.481} \\
         PatchTST & 31.17 & 183.6 & 1.530 & 0.469\\
         TimesNet & 34.35 & 191.9 & 1.604 & 0.415 \\
         Crossformer & 39.58 & 193.6 & 2.141 & 0.394 \\
         % Transformer & 34.81 & 192.7 & 1.601 & 0.413\\ 
         \hline
    \end{tabular}
    \caption{Test performance of the deep learning models. The best results are in bold.}
    \label{tab:test_results}
\end{table}

\begin{figure}[htbp]
\includegraphics[width=0.50\textwidth]{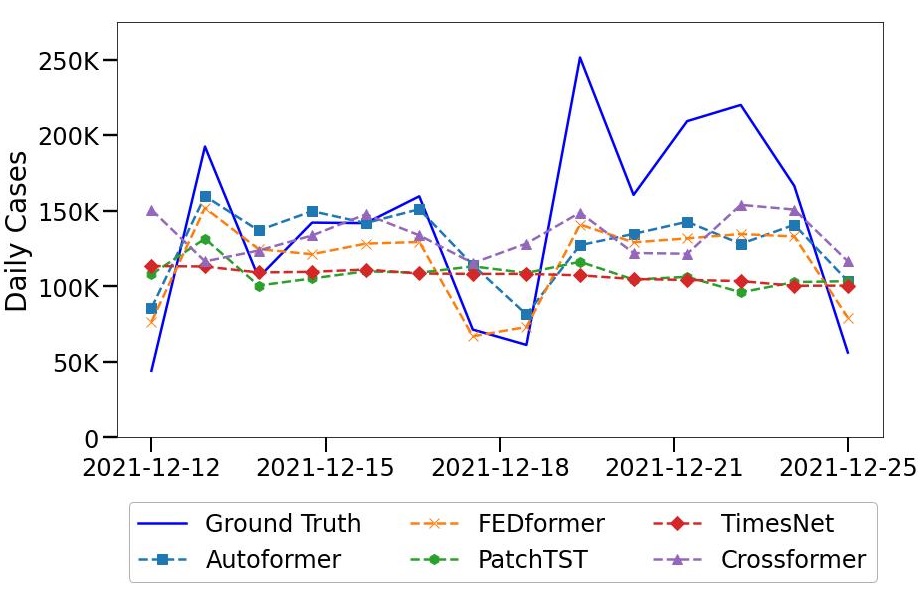}
\caption{Test predictions comparison with ground truth aggregated over all counties.}
\label{fig:test_comparison}
\end{figure}

Table \ref{tab:test_results} shows the test results of the models. The FEDformer model performs best with the lowest MAE, RMSE, RMSLE loss, and highest R$^2$-score. We have used this FEDformer model in our later section for the interpretation tasks. Figure \ref{fig:test_comparison} plots the ground truth and model predictions aggregated over the 3,142 US counties. The aggregated plot is shown for simplicity. However, the prediction and evaluation are done at each US county level. 

\begin{table}[htbp]
    \centering
    \resizebox{8.5cm}{!}{
    \begin{tabular}{|p{2.9cm}|c|c|c|c|} \hline
         \multirow{2}{=}{\textbf{Method}}& \multicolumn{2}{|c|}{Comprehensiveness ($\uparrow$)} & \multicolumn{2}{|c|}{Sufficiency($\downarrow$)} \\ \cline{2-5}
         & MAE & MSE & MAE & MSE \\ \hline
         Feature Ablation & 4.91 & 8.64 & 9.53 & 10.5\\
         Feature Permutation & 4.00 & 7.08 &  8.00 & 8.28\\
         Morris Sensitivity & \textbf{6.23} & 9.39 & \textbf{5.85} & \textbf{5.46}\\
         Feature Occlusion & 4.89 & 8.44 & 9.49 & 10.4 \\
         Augmented F.O. & 4.18 & 7.66 & 7.96 & 8.09 \\
         Deep Lift & 5.72 & \textbf{9.54} & 8.90 & 9.43\\
         Integrated Gradients & 5.52 & 9.09 & 9.25 & 10.2 \\
         Gradient Shap & 4.78 & 8.17  & 8.04 & 8.27 \\
         % Lime & 3.19 & 4.33 & 8.66 & 10.7 \\
         \hline
    \end{tabular}}
    \caption{AOPCR results of the interpretation using the FEDformer model on the \textbf{COVID} test set.}
    \label{tab:int_results_no_ground_truth}
\end{table}

\subsection{Two Additional Experimental Setup}

The Electricity and Traffic datasets are divided into train, validation test sets using a 8:1:1 split. We arbitrarily chose Crossformer for these two datasets. The MAE and MSE errors on the test are 0.2534 and 0.1292 for the Electricity dataset. For the Traffic dataset, the MAE and MSE errors are 0.2938 and 0.2258 respectively.

\subsection{Interpretation Evaluation}
Following the same steps as described in Section \ref{sec:eval_without_truth}, we apply all of our eight interpretation methods on our \textbf{COVID} dataset and display our results in Table \ref{tab:int_results_no_ground_truth}. Furthermore, we showcase the performance of an arbitrarily chosen subset of interpretation methods in Table \ref{tab:int_results_electricity} and \ref{tab:int_results_traffic} for the Electricity and Traffic datasets.

\begin{table}[htbp]
    \centering
    \resizebox{8.5cm}{!}{
    \begin{tabular}{|p{3cm}|c|c|c|c|} \hline
         \multirow{2}{=}{\textbf{Method}}& \multicolumn{2}{|c|}{Comprehensiveness ($\uparrow$)} & \multicolumn{2}{|c|}{Sufficiency($\downarrow$)} \\ \cline{2-5}
         & MAE & MSE & MAE & MSE \\ \hline
         Feature Ablation & \textbf{13.4} & \textbf{12.2} & 17.0 & 18.1\\
         Feature Permutation & 7.57 & 5.28 &  \textbf{15.2} & \textbf{14.8}\\
         Feature Occlusion & 13.3 & 12.2 & 17.1 & 18.4  \\
         Augmented F.O. & 8.27 & 6.12 & 15.3 & 15.1 \\
         \hline
    \end{tabular}}
    \caption{AOPCR results of the interpretation using the Crossformer model on the \textbf{Electricity} test set.}
    \label{tab:int_results_electricity}
\end{table}

\begin{table}[htbp]
    \centering
    \resizebox{8.5cm}{!}{
    \begin{tabular}{|p{3cm}|c|c|c|c|} \hline
         \multirow{2}{=}{\textbf{Method}}& \multicolumn{2}{|c|}{Comprehensiveness ($\uparrow$)} & \multicolumn{2}{|c|}{Sufficiency($\downarrow$)} \\ \cline{2-5}
         & MAE & MSE & MAE & MSE \\ \hline
         Feature Ablation & 10.8 & 7.70 & \textbf{16.3} & \textbf{16.6}\\
         Feature Permutation & 8.20 & 4.72 &  19.6 & 22.1 \\
         Feature Occlusion & \textbf{10.9} & \textbf{7.76} & 16.4 & 16.8  \\
         Augmented F.O. & 8.39 & 4.89 & 19.4 & 21.8 \\
         \hline
    \end{tabular}}
    \caption{AOPCR results of the interpretation using the Crossformer model on the \textbf{Traffic} test set.}
    \label{tab:int_results_traffic}
\end{table}

\section{Evaluating Age Group Sensitivity with Ground Truth}
\label{sec:ground_truth}
This section presents an innovative way to evaluate the importance of age groups from CDC \cite{weekly_cases_by_age}. The previous section uses performance drop-based methods to evaluate interpretation because of the lack of ground truth for interpretation. And that evaluation is done at each county and daily level, the same as the prediction task. 

\begin{figure}[htbp]
\includegraphics[width=0.46\textwidth]{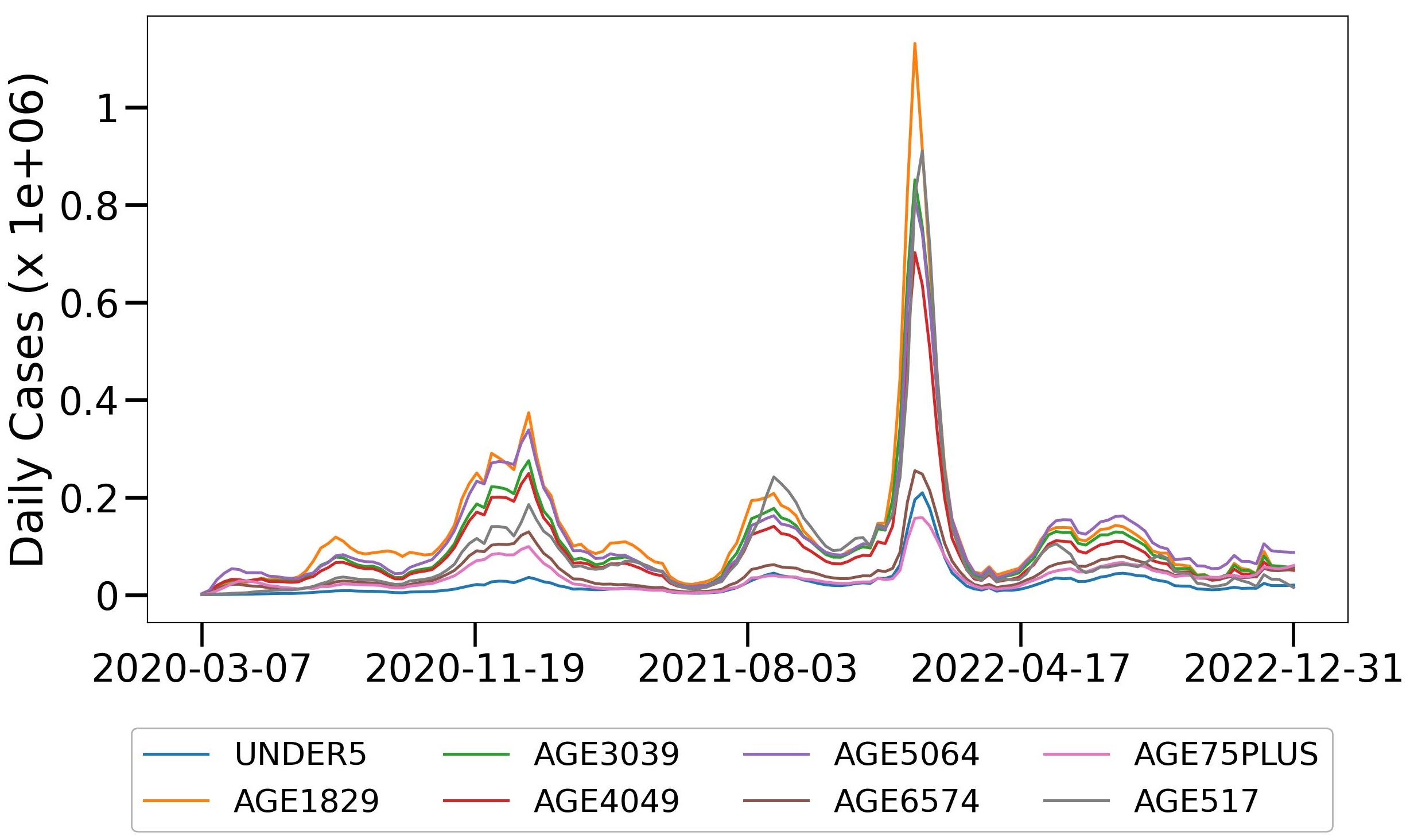}
\caption{Weekly COVID-19 cases \cite{weekly_cases} for each of the eight age subgroups over the study period .}
\label{fig:weekly_ground_truth}
\end{figure}

However, the COVID-19 cases by age groups from CDC \cite{weekly_cases_by_age} comes at a weekly rate for the whole United States. Hence can only be evaluated at the weekly and aggregated to the country level. Figure \ref{fig:weekly_ground_truth} shows the ground truth values by age groups over our whole dataset. Table \ref{tab:weekly_ground_truth} shows the summary and rank of different age groups for the test period. 

\begin{table}[ht!]
    \centering
    \begin{tabular}{|c|c|c|c|} \hline
        \textbf{Age Group} & \multicolumn{3}{|c|}{\textbf{Total Cases}} \\ \cline{2-4}
         & \textbf{Actual} & \textbf{Normalized (\%)} & \textbf{Rank} \\ \hline
         $<$ 5 & 99654 & 3.569 & 7\\
        5-17 & 404420 & 14.48 & 4\\
        18-29 & 686648 & 24.59 & 1\\
         30-39 & 539684 & 19.32 & 2\\
         40-49 & 393727 & 14.10 & 5\\
         50-64 & 443701 & 15.89 & 3\\
         65-74 & 141490 & 5.067 & 6\\
         75+ & 83086 & 2.975 & 8\\ \hline
    \end{tabular}
    \caption{COVID-19 cases in all US counties by age groups during the test period, 12-25 Dec 2021.}
    \label{tab:weekly_ground_truth}
\end{table}

The weekly rank of the age groups by infection rate doesn't change much with time. However, the infection rate itself can vary a lot and be more challenging. Hence we focus on predicting the normalized (l1-norm) infection rate for each age group. The normalization is done to understand how each age group contributes to the overall infection spread each week.

\subsection{Evaluation on Extended Dataset}
\begin{figure}[ht!]
\includegraphics[width=0.51\textwidth]{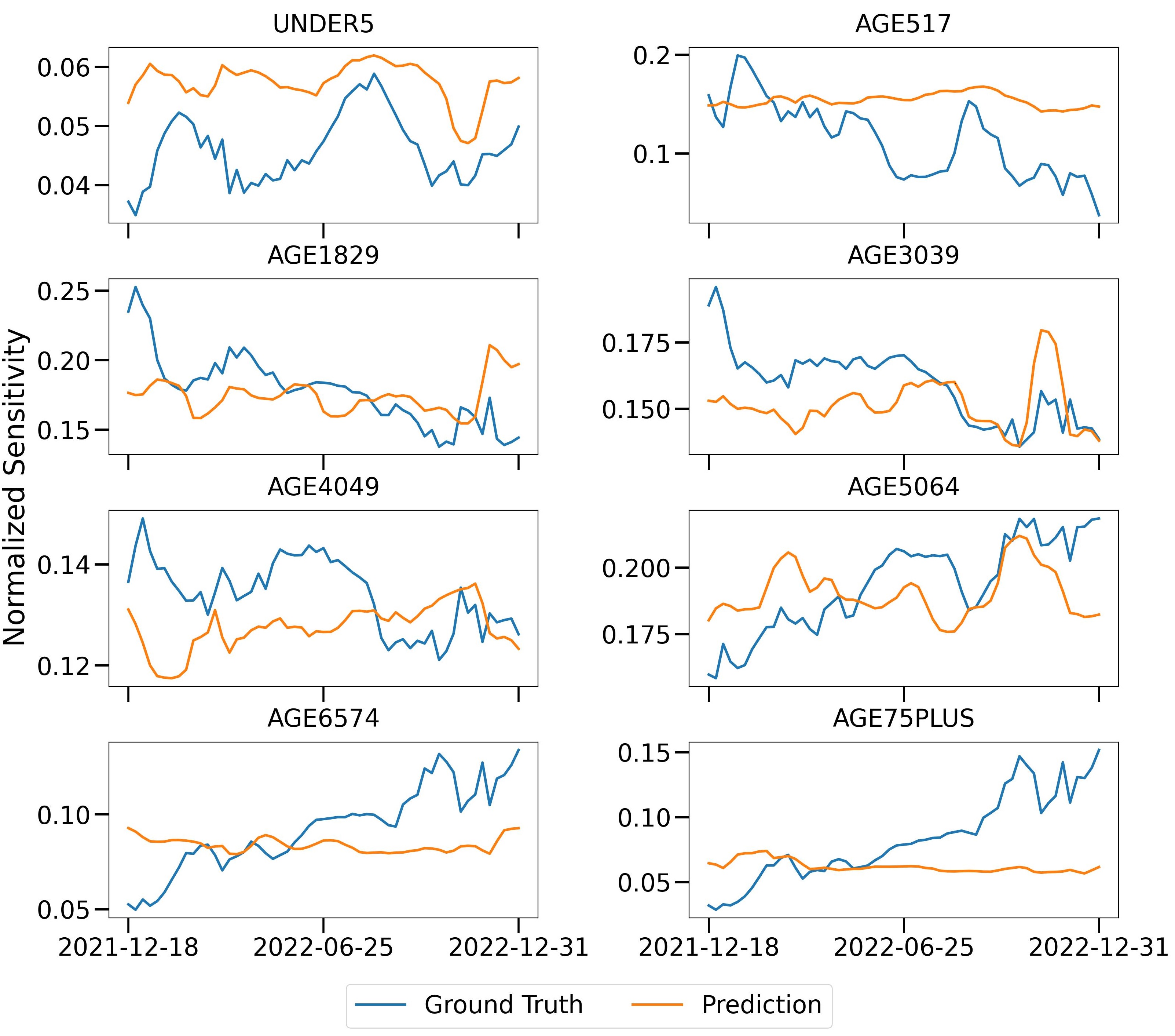}
\caption{Predicted age sensitivity on the extended dataset with weekly COVID-19 cases by age groups as ground truth.}
\label{fig:attribution_updated_ages}
\end{figure}

Our test set initially comprises only two weeks of data, but we extend the evaluation until December 31, 2022, encompassing over a year's worth of test data. This extension aims to demonstrate the alignment of our predicted age sensitivity with the actual cases reported by the CDC for the United States. Figure \ref{fig:attribution_updated_ages} compares the extended dataset. Both predicted attribution from the importance matrix 
$\phi$ and actual sensitivity (cases by age groups) are normalized to sum to 1.00. The results show that the trends of different age groups keep changing with time. 

\subsection{Interpreting different feature attribution}
Table \ref{tab:feature_importance} shows the aggregated importance of the input features over the test set using the Morris Sensitivity method. We average the attribution matrix $\phi$ and normalize the scores to percentages. The past COVID-19 cases are the most important. Among other features, the age groups 18-29 and 65-75 are more important.  
\begin{table}[hbp!]
    \centering
    \begin{tabular}{|c|c|c|c|} \hline
        \textbf{Feature} & \textbf{Importance} & \textbf{Feature} & \textbf{Importance}\\ \hline
        UNDER5 & 4.737 & AGE75PLUS & 4.695 \\
        AGE517 & 4.785 & Vaccination & 4.385 \\
        AGE1829 & 5.264 & Cases  & \textbf{41.98}\\
        AGE3039 & 4.946 & Day & 4.961  \\
        AGE4049 & 5.191 & Month & 4.611  \\ 
        AGE5064 & 4.594 & Weekday & 4.594 \\
        AGE6575 & 5.258  & & \\
         \hline
    \end{tabular}
    \caption{Feature importance (\%) evaluated on the test set by aggregating attribution scores for the input features.}
    \label{tab:feature_importance}
\end{table}

\begin{figure}[ht!]
\includegraphics[width=0.50\textwidth]{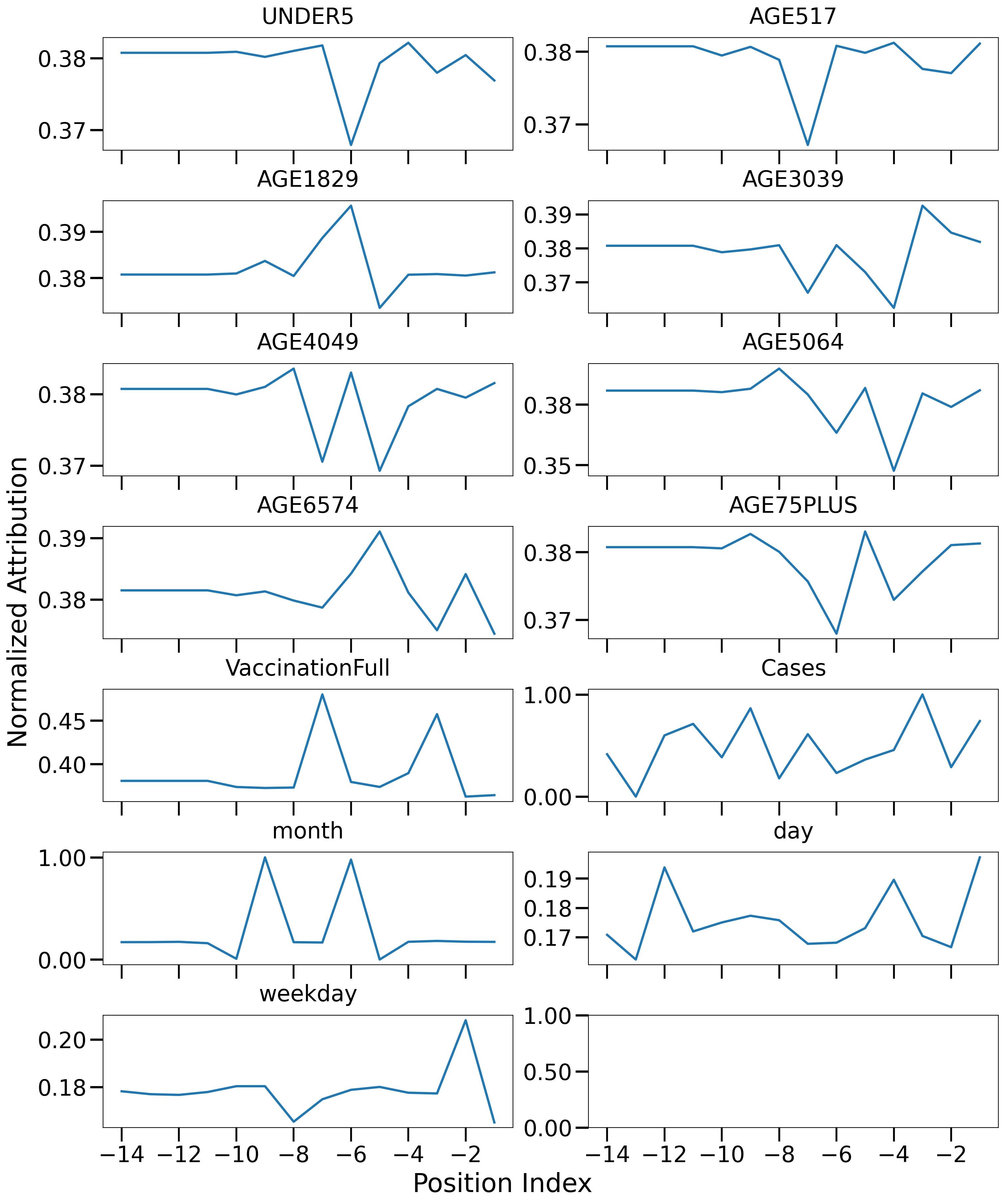}
\caption{Interpreting different feature attribution for the Los Angeles, California county from our test set using the FEDformer model and Feature Ablation method.}
\label{fig:attribution_test_6037}
\end{figure}

\textbf{An interpretation example} is shown in Figure \ref{fig:attribution_test_6037} for the Los Angeles, California county from our test period. The position index from -14 to -1 is for the input, while position 0 to 13 is for the prediction horizon. The result shows that the same feature from different lookback positions has different impact on the predictions. The past cases are most important and working-age populations (AGE3039 and AGE4049) also get higher importance. Recent cases get higher attribution values, showing the model's prediction is more relevant to recent infection rates.

We calculated how much this normalized infection rate differs from the predicted scores using different interpretation methods. The importance matrix $\phi \in \mathbf{R}^{\tau_{max} \times J \times L}$ is aggregated over the lookback window $L$ to find the overall impact of each age feature $j$ for the prediction horizon $\tau$.  The difference between the actual and the predicted rate through interpretation is evaluated using MAE, RMSE, and NDCG (Normalized Discounted Cumulative Gain). The NDCG ranking metric returns a high value if true labels are ranked high by the predicted scores. 

Table \ref{tab:results_weekly} shows the final results. There is no single best method from the results but most methods achieve high accuracy in predicting the age group sensitivities.

\begin{table}[htbp]
    \centering
    \begin{tabular}{|p{2.9cm}|c|c|c|} \hline
        \multirow{2}{*}{\textbf{Method}} & \multicolumn{3}{|c|}{\textbf{Metrics}} \\ \cline{2-4}
         & \textbf{MAE} $\downarrow$& \textbf{RMSE} $\downarrow$& \textbf{NDCG} $\uparrow$\\ \hline
Feature Ablation &\textbf{ 0.0336 }& 0.0426 & 0.9849 \\
Feature Permutation & 0.0339 & 0.0438 & \textbf{0.9900} \\
Morris Sensitivity & 0.0350 & \textbf{0.0393} & 0.9587 \\
Feature Occlusion & 0.0338 & 0.0429 & 0.9851 \\
Augmented F.O. & 0.0346 & 0.0446 & 0.9587 \\
Deep Lift & 0.0383 & 0.0455 & 0.9641 \\
Integrated Gradients & 0.0386 & 0.0460 & 0.9641 \\
Gradient Shap & 0.0336 & 0.0431 & 0.9713 \\
         \hline
    \end{tabular}
    \caption{Evaluation of predicted normalized attribution scores with normalized weekly \textbf{COVID-19 cases} by age group in the test period, 12-25 Dec 2021. The best results are in bold.}
    \label{tab:results_weekly}
\end{table}

\section{Discussion \label{sec:discussion}}
We discuss the time complexity of our approach and the code reproducibility of our work. 

\begin{table}[hbtp]
    \centering
    \begin{tabular}{|p{2.9cm}|c|p{2.4cm}|c|} \hline
        \textbf{Method} & \textbf{Time} & \textbf{Method} & \textbf{Time} \\ \hline
Morris Sensitivity & 294.1 & Feature Ablation & 162.9 \\
Feature Permutation & 162.8 & Augmented F.O. & 165.3 \\
Feature Occlusion & 165.3 & Deep Lift & 55.40 \\
Integrated Gradients & 120.4 & Gradient Shap  & 59.58 \\
         \hline
    \end{tabular}
    \caption{Execution time (seconds) of the interpretation methods on the test set.}
    \label{tab:time_complexity}
\end{table}

\subsection{Time Complexity}
Execution time is important for real-time applications. We report the execution time of our interpretation methods and experiments on the test set in Table \ref{tab:time_complexity}. We observe the gradients-based methods perform faster. The Morris Sensitivity method is slower due to multiple sampling of the input features.

\subsection{Code Reproducibility }
Our code and datasets are publicly available on GitHub at https://github.com/UVA-MLSys/COVID-19-age-groups. We have also published a singularity container documenting the software versions. This also helps to readily deploy it on any HPC cluster. Our random methods are seeded to ensure reproducibility.

\section{Conclusion and Future Work}\label{sec:conclusion}

In this work, we showed how to interpret state-of-the-art time series transformer models using very dynamic and complex COVID-19 infection data and two benchmark time series datasets using Electricity and Traffic. We provide a thorough analysis of recent times series interpretation methods performance on the state-of-the-art Transformer models. Our results show that we can not only interpret changes in feature importance over past time steps but also predict the sensitivity of these features in future horizons. Our proposed framework helps us understand the impacts of past observations, but also predict their impacts in the future. Future works include capturing higher-order relations between the input features, understanding spatiotemporal interpretations better, and benchmarking more time series domains with our framework. 

\section{Acknowledgment}\label{sec: acknowledgement} This work is partially supported by NSF grant CCF-1918626  Expeditions: Collaborative  Research: Global Pervasive Computational Epidemiology, and NSF Grant 1835631 for CINES: A Scalable Cyberinfrastructure for Sustained Innovation in Network Engineering and Science.

\bibliography{aaai24}

\begin{thebibliography}{32}
\providecommand{\natexlab}[1]{#1}

\bibitem[{Amann et~al.(2020)Amann, Blasimme, Vayena, Frey, and Madai}]{Amann2020ExplainabilityFA}
Amann, J.; Blasimme, A.; Vayena, E.; Frey, D.; and Madai, V.~I. 2020.
\newblock Explainability for artificial intelligence in healthcare: a multidisciplinary perspective.
\newblock \emph{BMC Medical Informatics and Decision Making}, 20.

\bibitem[{Arik, Yoder, and Pfister(2022)}]{arik2022self}
Arik, S.~O.; Yoder, N.~C.; and Pfister, T. 2022.
\newblock Self-Adaptive Forecasting for Improved Deep Learning on Non-Stationary Time-Series.
\newblock \emph{arXiv preprint arXiv:2202.02403}.

\bibitem[{{Centers for Disease Control and Prevention}(2023{\natexlab{a}})}]{weekly_cases}
{Centers for Disease Control and Prevention}. 2023{\natexlab{a}}.
\newblock {COVID-19 Weekly Cases and Deaths by Age, Race/Ethnicity, and Sex}.

\bibitem[{{Centers for Disease Control and Prevention}(2023{\natexlab{b}})}]{weekly_cases_by_age}
{Centers for Disease Control and Prevention}. 2023{\natexlab{b}}.
\newblock {COVID-19 Weekly Cases and Deaths by Age, Race/Ethnicity, and Sex}.

\bibitem[{Clement et~al.(2021)Clement, Ponnusamy, Sriharipriya, and Nandakumar}]{clement2021survey}
Clement, J.~C.; Ponnusamy, V.; Sriharipriya, K.; and Nandakumar, R. 2021.
\newblock A survey on mathematical, machine learning and deep learning models for COVID-19 transmission and diagnosis.
\newblock \emph{IEEE reviews in biomedical engineering}, 15: 325--340.

\bibitem[{DeYoung et~al.(2019)DeYoung, Jain, Rajani, Lehman, Xiong, Socher, and Wallace}]{deyoung2019eraser}
DeYoung, J.; Jain, S.; Rajani, N.~F.; Lehman, E.; Xiong, C.; Socher, R.; and Wallace, B.~C. 2019.
\newblock ERASER: A benchmark to evaluate rationalized NLP models.
\newblock \emph{arXiv preprint arXiv:1911.03429}.

\bibitem[{Doshi-Velez and Kim(2017)}]{doshi2017towards}
Doshi-Velez, F.; and Kim, B. 2017.
\newblock Towards a rigorous science of interpretable machine learning.
\newblock \emph{arXiv preprint arXiv:1702.08608}.

\bibitem[{Enguehard(2023)}]{enguehard2023time}
Enguehard, J. 2023.
\newblock Time Interpret: a Unified Model Interpretability Library for Time Series.
\newblock \emph{arXiv preprint arXiv:2306.02968}.

\bibitem[{Erion et~al.(2019)Erion, Janizek, Sturmfels, Lundberg, and Lee}]{Erion2019LearningEM}
Erion, G.~G.; Janizek, J.~D.; Sturmfels, P.; Lundberg, S.~M.; and Lee, S.-I. 2019.
\newblock Learning Explainable Models Using Attribution Priors.
\newblock \emph{ArXiv}, abs/1906.10670.

\bibitem[{Ismail et~al.(2020)Ismail, Gunady, Corrada~Bravo, and Feizi}]{ismail2020benchmarking}
Ismail, A.~A.; Gunady, M.; Corrada~Bravo, H.; and Feizi, S. 2020.
\newblock Benchmarking deep learning interpretability in time series predictions.
\newblock \emph{Advances in neural information processing systems}, 33: 6441--6452.

\bibitem[{Iwanaga, Usher, and Herman(2022)}]{Iwanaga2022}
Iwanaga, T.; Usher, W.; and Herman, J. 2022.
\newblock Toward {SALib} 2.0: {Advancing} the accessibility and interpretability of global sensitivity analyses.
\newblock \emph{Socio-Environmental Systems Modelling}, 4: 18155.

\bibitem[{Kim et~al.(2022)Kim, Min, Nam, Song, Yoon, Kim, and Lee}]{kim2022covid}
Kim, D.; Min, H.; Nam, Y.; Song, H.; Yoon, S.; Kim, M.; and Lee, J.-G. 2022.
\newblock Covid-eenet: Predicting fine-grained impact of COVID-19 on local economies.
\newblock In \emph{Proceedings of the AAAI Conference on Artificial Intelligence}, volume~36, 11971--11981.

\bibitem[{Kokhlikyan et~al.(2020)Kokhlikyan, Miglani, Martin, Wang, Alsallakh, Reynolds, Melnikov, Kliushkina, Araya, Yan, and Reblitz-Richardson}]{kokhlikyan2020captum}
Kokhlikyan, N.; Miglani, V.; Martin, M.; Wang, E.; Alsallakh, B.; Reynolds, J.; Melnikov, A.; Kliushkina, N.; Araya, C.; Yan, S.; and Reblitz-Richardson, O. 2020.
\newblock Captum: A unified and generic model interpretability library for PyTorch.
\newblock arXiv:2009.07896.

\bibitem[{Molnar(2020)}]{molnar2020interpretable}
Molnar, C. 2020.
\newblock \emph{Interpretable machine learning}.
\newblock Lulu. com.

\bibitem[{Morris(1991)}]{morris1991factorial}
Morris, M. 1991.
\newblock Factorial sampling plans for preliminary computational experiments.
\newblock \emph{Technometrics}, 33(2): 161--174.

\bibitem[{Nie et~al.(2023)Nie, Nguyen, Sinthong, and Kalagnanam}]{nie2023patchtst}
Nie, Y.; Nguyen, N.~H.; Sinthong, P.; and Kalagnanam, J. 2023.
\newblock A time series is worth 64 words: Long-term forecasting with transformers.
\newblock \emph{International Conference on Learning Representations}.

\bibitem[{Ozyegen, Ilic, and Cevik(2022)}]{ozyegen2022evaluation}
Ozyegen, O.; Ilic, I.; and Cevik, M. 2022.
\newblock Evaluation of interpretability methods for multivariate time series forecasting.
\newblock \emph{Applied Intelligence}, 1--17.

\bibitem[{Ramchandani, Fan, and Mostafavi(2020)}]{ramchandani2020deepcovidnet}
Ramchandani, A.; Fan, C.; and Mostafavi, A. 2020.
\newblock Deepcovidnet: An interpretable deep learning model for predictive surveillance of covid-19 using heterogeneous features and their interactions.
\newblock \emph{Ieee Access}, 8: 159915--159930.

\bibitem[{Rodriguez et~al.(2021)Rodriguez, Tabassum, Cui, Xie, Ho, Agarwal, Adhikari, and Prakash}]{rodriguez2021deepcovid}
Rodriguez, A.; Tabassum, A.; Cui, J.; Xie, J.; Ho, J.; Agarwal, P.; Adhikari, B.; and Prakash, B.~A. 2021.
\newblock Deepcovid: An operational deep learning-driven framework for explainable real-time covid-19 forecasting.
\newblock In \emph{Proceedings of the AAAI Conference on Artificial Intelligence}, volume~35, 15393--15400.

\bibitem[{Rojat et~al.(2021)Rojat, Puget, Filliat, Del~Ser, Gelin, and D{\'\i}az-Rodr{\'\i}guez}]{rojat2021explainable}
Rojat, T.; Puget, R.; Filliat, D.; Del~Ser, J.; Gelin, R.; and D{\'\i}az-Rodr{\'\i}guez, N. 2021.
\newblock Explainable artificial intelligence (xai) on timeseries data: A survey.
\newblock \emph{arXiv preprint arXiv:2104.00950}.

\bibitem[{Shrikumar, Greenside, and Kundaje(2017)}]{shrikumar2017learning}
Shrikumar, A.; Greenside, P.; and Kundaje, A. 2017.
\newblock Learning important features through propagating activation differences.
\newblock In \emph{International conference on machine learning}, 3145--3153. PMLR.

\bibitem[{Sundararajan, Taly, and Yan(2017)}]{Sundararajan2017AxiomaticAF}
Sundararajan, M.; Taly, A.; and Yan, Q. 2017.
\newblock Axiomatic Attribution for Deep Networks.
\newblock In \emph{International Conference on Machine Learning}.

\bibitem[{Suresh et~al.(2017)Suresh, Hunt, Johnson, Celi, Szolovits, and Ghassemi}]{Suresh2017ClinicalIP}
Suresh, H.; Hunt, N.; Johnson, A. E.~W.; Celi, L.~A.; Szolovits, P.; and Ghassemi, M. 2017.
\newblock Clinical Intervention Prediction and Understanding using Deep Networks.
\newblock \emph{ArXiv}, abs/1705.08498.

\bibitem[{Turb{\'e} et~al.(2023)Turb{\'e}, Bjelogrlic, Lovis, and Mengaldo}]{turbe2023evaluation}
Turb{\'e}, H.; Bjelogrlic, M.; Lovis, C.; and Mengaldo, G. 2023.
\newblock Evaluation of post-hoc interpretability methods in time-series classification.
\newblock \emph{Nature Machine Intelligence}, 5(3): 250--260.

\bibitem[{{US Census Bureau}(2020)}]{pop2020}
{US Census Bureau}. 2020.
\newblock County Population by Characteristics: 2010-2020.

\bibitem[{Wu et~al.(2023{\natexlab{a}})Wu, Hu, Liu, Zhou, Wang, and Long}]{wu2023timesnet}
Wu, H.; Hu, T.; Liu, Y.; Zhou, H.; Wang, J.; and Long, M. 2023{\natexlab{a}}.
\newblock TimesNet: Temporal 2D-Variation Modeling for General Time Series Analysis.
\newblock In \emph{International Conference on Learning Representations}.

\bibitem[{Wu et~al.(2023{\natexlab{b}})Wu, Hu, Liu, Zhou, Wang, and Long}]{TSlib}
Wu, H.; Hu, T.; Liu, Y.; Zhou, H.; Wang, J.; and Long, M. 2023{\natexlab{b}}.
\newblock TimesNet: Temporal 2D-Variation Modeling for General Time Series Analysis.
\newblock In \emph{International Conference on Learning Representations}.

\bibitem[{Wu et~al.(2021)Wu, Xu, Wang, and Long}]{wu2021autoformer}
Wu, H.; Xu, J.; Wang, J.; and Long, M. 2021.
\newblock Autoformer: Decomposition Transformers with {Auto-Correlation} for Long-Term Series Forecasting.
\newblock In \emph{Advances in Neural Information Processing Systems}.

\bibitem[{Zeiler and Fergus(2013)}]{Zeiler2013VisualizingAU}
Zeiler, M.~D.; and Fergus, R. 2013.
\newblock Visualizing and Understanding Convolutional Networks.
\newblock In \emph{European Conference on Computer Vision}.

\bibitem[{Zhang and Yan(2022)}]{zhang2022crossformer}
Zhang, Y.; and Yan, J. 2022.
\newblock Crossformer: Transformer utilizing cross-dimension dependency for multivariate time series forecasting.
\newblock In \emph{The Eleventh International Conference on Learning Representations}.

\bibitem[{Zhao et~al.(2023)Zhao, Shi, Wu, Yang, Song, and Liu}]{zhao2023interpretation}
Zhao, Z.; Shi, Y.; Wu, S.; Yang, F.; Song, W.; and Liu, N. 2023.
\newblock Interpretation of Time-Series Deep Models: A Survey.
\newblock \emph{arXiv preprint arXiv:2305.14582}.

\bibitem[{Zhou et~al.(2022)Zhou, Ma, Wen, Wang, Sun, and Jin}]{zhou2022fedformer}
Zhou, T.; Ma, Z.; Wen, Q.; Wang, X.; Sun, L.; and Jin, R. 2022.
\newblock {FEDformer}: Frequency enhanced decomposed transformer for long-term series forecasting.
\newblock In \emph{Proc. 39th International Conference on Machine Learning (ICML 2022)}.

\end{thebibliography}
%\appendix
\end{document}